\documentclass[letterpaper, 10 pt, conference]{ieeeconf}  %

\IEEEoverridecommandlockouts                              %

\usepackage{graphics} %
\usepackage{epsfig} %
\usepackage{amsmath} %
\usepackage{amssymb}  %
\usepackage{lipsum}
\usepackage[dvipsnames]{xcolor}
\usepackage{hyperref}
\usepackage{booktabs}
\usepackage{algorithm}
\usepackage{algpseudocode}
\usepackage{url}
\usepackage{cite}

\newcommand{\argmax}{\mathop{\mathrm{\arg\!\max}}}

\title{\LARGE \bf
RoboHop: Segment-based Topological Map Representation for Open-World Visual Navigation
}

\author{Sourav Garg$^{1}$,  Krishan Rana$^{*2}$, Mehdi Hosseinzadeh$^{*1}$, Lachlan Mares$^{*1}$, \\Niko S\"underhauf$^{2}$, Feras Dayoub$^{1}$, Ian Reid$^{1,3}$%
\thanks{$^{1}$ Australian Institute for Machine Learning (AIML), The University of Adelaide, Australia.
        }%
\thanks{$^{2}$ Queensland University of Technology, Australia.
        }%
\thanks{$^{3}$ Mohamed Bin Zayed University of Artificial Intelligence, UAE.
        }%
\thanks{*Equal Contribution}%
}

\begin{document}

\maketitle
\thispagestyle{empty}
\pagestyle{empty}

\begin{abstract}

Mapping is crucial for spatial reasoning, planning and robot navigation. Existing approaches range from metric, which require precise geometry-based optimization, to purely topological, where image-as-node based graphs lack explicit object-level reasoning and interconnectivity. In this paper, we propose a novel topological representation of an environment based on \textit{image segments}, which are semantically meaningful and open-vocabulary queryable, conferring several advantages over previous works based on pixel-level features. Unlike 3D scene graphs, we create a purely topological graph with segments as nodes, where edges are formed by \textit{a)} associating segment-level descriptors between pairs of consecutive images and \textit{b)} connecting neighboring segments within an image using their pixel centroids. This unveils a \textit{continuous sense of a place}, defined by inter-image persistence of segments along with their intra-image neighbours. It further enables us to represent and update segment-level descriptors through neighborhood aggregation using graph convolution layers, which improves robot localization based on segment-level retrieval. 
Using real-world data, we show how our proposed map representation can be used to \textit{i)} generate navigation plans in the form of \textit{hops over segments} and \textit{ii)} search for target objects using natural language queries describing spatial relations of objects. Furthermore, we quantitatively analyze data association at the segment level, which underpins inter-image connectivity during mapping and segment-level localization when revisiting the same place. Finally, we show preliminary trials on segment-level `hopping' based zero-shot real-world navigation. Project page with supplementary details: \url{oravus.github.io/RoboHop/}.

\end{abstract}

\section{INTRODUCTION}

A map of an environment represents spatial understanding which an embodied agent can use to operate in that environment. This manifests in existing approaches in multiple ways, e.g., 3D metric maps used for precise operations~\cite{conceptfusion,sarlin2022lamar}, implicit maps as a robot's memory~\cite{wu2023daydreamer}, hierarchical 3DSGs based explicit memory~\cite{ravichandran2022hierarchical}, and topological maps with image-level connectivity for robot navigation~\cite{SPTM2018semi,shah2022lmnav,chen2019behavioral,chaplot2020neural}. Metric maps enable direct spatial reasoning, e.g.,  6-DoF poses of a driverless vehicle, or measuring distances to or between physical entities in the environment. Even for purely topological representations, some spatial reasoning can be encoded through image-level connectivity, e.g., recent advances in bio-inspired topological navigation~\cite{SPTM2018semi} and the follow-up work~\cite{shah2022lmnav,li2020learning,meng2020scaling}. However, such topological representations discretized by images are limited in their semantic expressivity as the physical entities in the world are never explicitly represented or associated across images.

\begin{figure}
    \centering
    \includegraphics[width=0.4\textwidth]{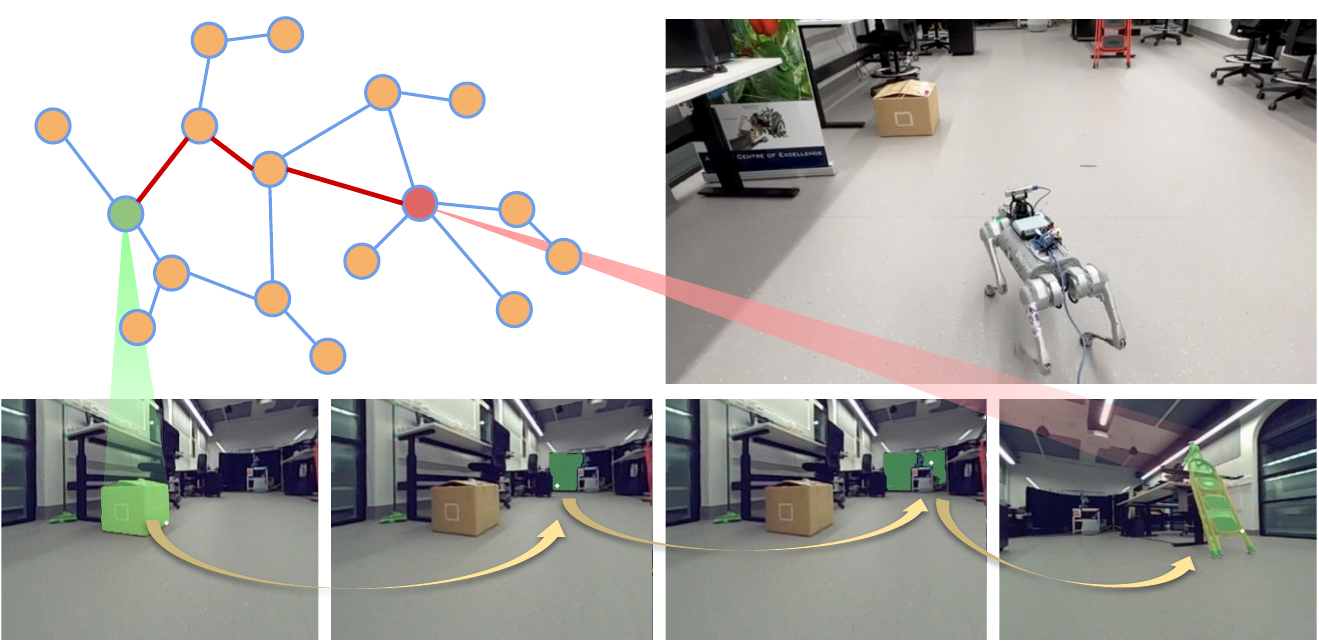}
    \footnotesize{Segment-level Plan to Navigate  from \texttt{Cardboard Box} to \texttt{Ladder}.}
    \caption{We present a topological, segment-based map representation which can generate navigation plans from open-vocabulary queries in the form of `hops' over segments to reach the goal, without needing a learned policy.}
    \label{fig:teaser}
\end{figure}

In this paper, we propose a novel topological representation of an environment based on \textit{image segments}.  
Unlike the use of pixel-level features~\cite{johns2011global}, the segments we use are semantically meaningful and 
open-vocabulary queryable. Our segments-based approach is enabled by recent advances in image segmentation, i.e., SAM~\cite{kirillov2023segment} and vision-language coupling, i.e., CLIP~\cite{radford2021learning}. We create a topological graph using image segments as nodes, with edges formed by \textit{a)} associating image segments within a temporal window of image observations and \textit{b)} connecting neighboring segments within an image using their pixel centroids.

We show how our map representation can be used to create intra-image \textit{hops} over inter-image \textit{segment tracks} to generate navigation plans and actions, as shown in Figure~\ref{fig:teaser}. Unlike existing image-level topological navigation methods~\cite{SPTM2018semi,shah2022lmnav,chen2019behavioral}, the use of segments directly enables finer-grained plan generation for object-goal navigation. 
Furthermore, we show how our proposed segment-level inter- and intra-image connectivity unveils a continuous sense of a `place'~\cite{garg2021your}, represented by a segment descriptor and its neighboring nodes. 
These segment descriptors are updated, enhanced and augmented with their neighbours via graph convolution. This rich descriptor enables accurate robot localization via segment-level retrieval.

In summary, the contributions of this paper are as follows: \textit{a)} We introduce a novel topological representation of environments, utilizing \textit{image segments} as nodes; this enables semantically rich and open-vocabulary queryable mapping. \textit{b)} We establish a novel mechanism for intra- and inter-image connectivity based on segment-level descriptors and pixel centroids. \textit{c)} We develop a unique method for generating semantically interpretable, segment-level plans for navigation, leveraging text-based queries for defining object-level source and target nodes. \textit{d)} We demonstrate the utility of our segment-level mapping, planning, and localization through preliminary trials of zero-shot real-world navigation.

\section{Related Work}

\textit{Mapping:} Mapping techniques fall into three main categories: 3D metric maps~\cite{murORB2,engel14eccv,schonberger2016structure,conceptfusion,dube2020segmap}, purely topological maps~\cite{FABMAP2010fab,SPTM2018semi}, and hybrid maps which often combine semantics with `topometric' information, e.g., 3D Scene Graphs~\cite{rosinol2021kimera,armeni20193d,kim20193,gay2019visual}. 3D approaches like ORB-SLAM~\cite{murORB2}, LSD-SLAM~\cite{engel14eccv}, and PTAM~\cite{PTAM2007} excel in accuracy but suffer from computational overhead and a lack of semantics, limiting their application in high-level task planning. Hybrid methods such as SLAM++~\cite{salas2013slam++} and QuadricSLAM~\cite{QuadricSLAM2019} attempt to address this by incorporating semantic information but remain computationally intensive.
Purely topological methods like FAB-MAP~\cite{FABMAP2010fab} and SPTM~\cite{SPTM2018semi} simplify the computational load by using graphs to represent places and paths but lack explicit object-level connectivity.

\textit{Navigation:} Semantic and spatial reasoning is crucial for 
object-goal navigation~\cite{anderson2018evaluation}, where a robot navigates toward a specified object represented through an image or a natural language instruction. Although some works have advocated for end-to-end learning through reinforcement~\cite{chen2018learning, wahid2021learning, bruce_sünderhauf_deepmind_london_deepmind_milford} or imitation~\cite{lee2021generalizable, pirlnav}, these approaches often necessitate large training datasets that are impractical in real-world scenarios. A less data-hungry alternative is to segregate the task into the classical three-step process: mapping, planning and then acting. Map-based strategies have exhibited superior modularity, scalability and interpretability, thus being suitable for real-world applications~\cite{chen2023train}. LM-Nav~\cite{shah2022lmnav} and TGSM~\cite{TSGM} build on SPTM~\cite{SPTM2018semi} to create topological graph representations, coupled with image-based CLIP features or closed-set object detections associated with each location. These representations can then be used to generate sub-goals which a robot can navigate towards with an image-based, low-level control policy. 
Learning such policies requires both environment- and embodiment-specific training data, limiting the generality of the approach. More recent work in this direction is aimed at creating foundation models for navigation~\cite{shah2023vint}. However, these topological maps with images-as-nodes lack explicit object-level reasoning, unless combined with 3D input~\cite{huang23vlmaps,TSGM,conceptgraphs}.    
In our work, we present a novel topological representation with `segments-as-nodes', which provides the robot with \textit{segment tracks of persistent entities}, where each node in the graph is connected to the next via segment matching across images. As segments disappear from parts of an image, other segments match to the next image allowing for a continuous \textit{hopping} over a stream of nodes. Such a representation enables a robot to progress towards a goal by ``segment servoing'' sub-goals, which relaxes the need for embodiment specific and sample-inefficient learned policies. Moreover, unlike existing image-based servoing~\cite{feng2021trajectory,bista2016appearance,mezouar2002path,hutchinson1996tutorial,cherubini2011visual,ahmadi2020visual,remazeilles20063d,diosi2011experimental,blanc2005indoor} and visual teach-and-repeat methods~\cite{furgale2010visual,vsegvic2009mapping,zhang2009robust,dall2021fast,mattamala2022efficient,krajnik2018navigation,halodova2019predictive,do2019high,krajnik2017image} for navigation, our map representation is purely topological \textit{and} based on segments~\cite{kirillov2023segment} which are semantically meaningful and open-vocabulary queryable.

\begin{figure*}
    \centering
    \includegraphics[width=\textwidth]{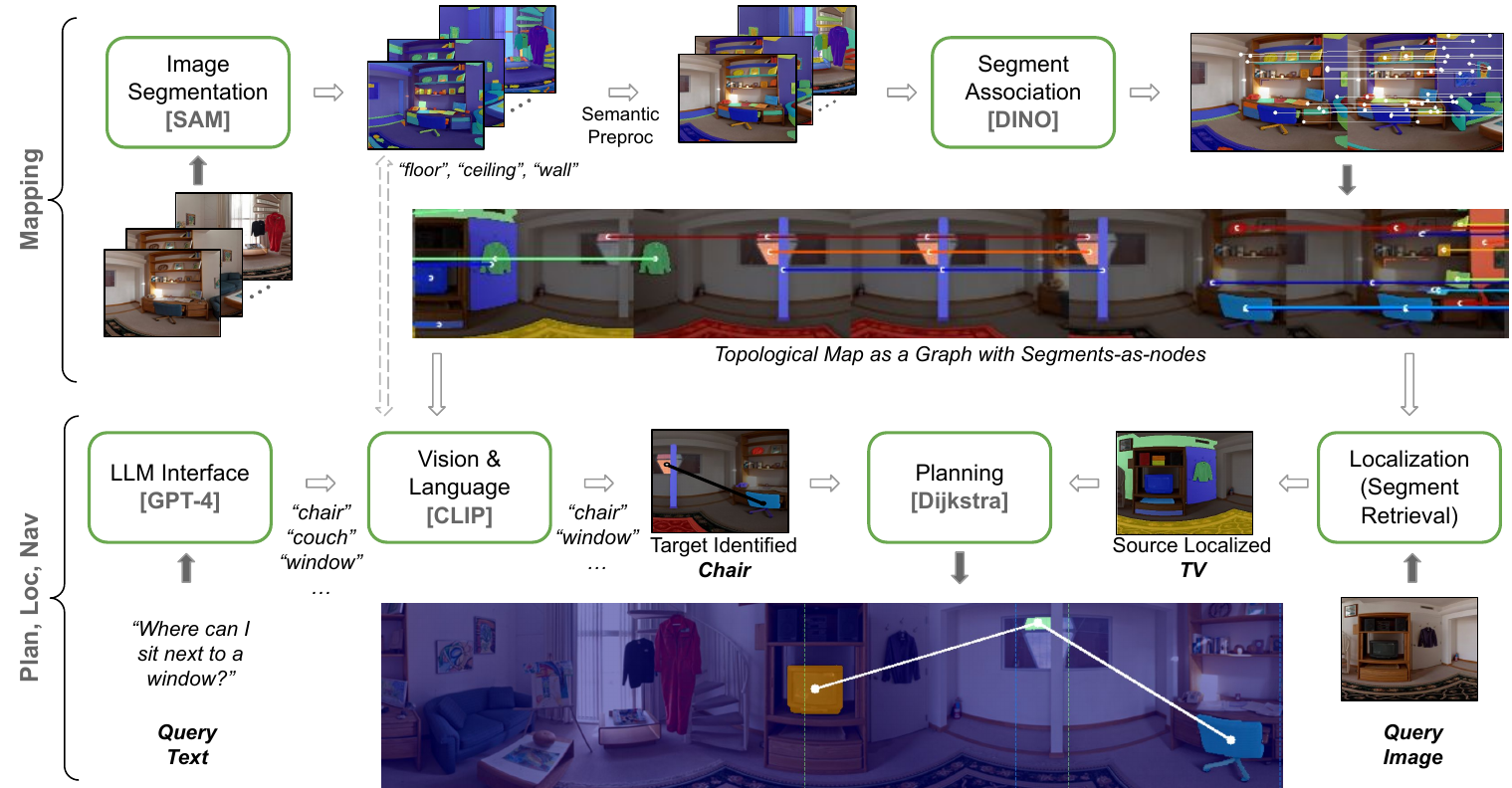}    
    \caption{Illustration of our overall pipeline from image segments to mapping, language querying, and planning.}
    \label{fig:pipeline}
\end{figure*}

\section{RoboHop}
Figure~\ref{fig:pipeline} illustrates our proposed pipeline for \textit{RoboHop} and its key modules: mapping, localization, planning, navigation and open-vocabulary natural language querying. 
\subsection{Mapping}\label{subsec:SegmentMap}
We define a map of an environment as a topological graph $\mathcal{G} = (\mathcal{N},\mathcal{E})$, where $\mathcal{N}$ and $\mathcal{E}$ represent the nodes and edges. For a given sequence of images $I^t \in I$, we first obtain image segmentation from a method such as SAM~\cite{kirillov2023segment}. The zero-shot capability of these recent foundation models is important because we do not want to tie our topological representation to a closed-world of known/recognised objects. Furthermore, these methods naturally support the link to richer descriptors and language models.

For each segment in an image, we define a node $n_i$ in $\mathcal{G}$ with attributes $(x_i,y_i,\textbf{M}_i,\textbf{h}_i^l)$. $(x_i, y_i)$ represent the pixel centroid of the binary mask $\textbf{M}_i$, $\textbf{h}_i^0$ represents the l2-normalized segment descriptor obtained by aggregating pixel-level deep features (using DINO~\cite{caron2021emerging} or DINOv2~\cite{oquab2023dinov2}) corresponding to $\textbf{M}_i$, and $l \in [0,l_{max}]$ is the layer index for descriptor aggregation in the graph (as explained later). As a semantic preprocessing step,  we also compute CLIP~\cite{radford2021learning} descriptors for individual segments (similar to~\cite{maalouf2023follow}) and exclude the segments with high (image-language) similarity to semantic labels for `stuff' (i.e., \textit{floor, ceiling, and wall}).

\subsubsection*{Edges} An edge $e_{ij}$ is defined as either of the two edge types: a) \textit{intra-image edges}, which are defined through the centroids of segments $(x_i^t, y_i^t)$ within each image $I^t$ using Delaunay Triangulation and b) \textit{inter-image edges}, which are defined through segment-level data association, i.e., vector dot product $s^{t,t'}_{ij} = \mathbf{h}_{i}^t \cdot \mathbf{h}_j^{t'}$ between node descriptors of an image pair $(I^t, I^{t'})$ as follows:
\begin{equation}
\mathcal{E}^{t,t'} = \{ (n_{i}^t, n_j^{t'}) \, | \, 
n_j^{t'} = \argmax_{k} s^{t,t'}_{ik}
\, \land \, s^{t,t'}_{ij} > \theta 
\}
\label{eq:edge}
\end{equation}
where $t'-t\in[1,3]$ and an edge between a pair of segment nodes $(n_i^t, n_j^{t'})$ only exists if $n_j^{t'}$ is the closest match for $n_i^t$ and their similarity is greater than a threshold $\theta$. If no edge is found for any segment in a particular image, we retain a single edge to its next image using the node pair with the highest similarity value. This ensures that our map is a connected graph. We do not define loop closure edges, which can be used to further enhance the map for shortcuts.

\subsubsection*{Node Descriptor \& Aggregation} 
The nodes in our map are based on segments which represent semantically meaningful entities in the environment. By defining a segment descriptor for each node based on robust features such as DINOv2~\cite{oquab2023dinov2} (e.g., see AnyLoc~\cite{keetha2023anyloc}), these segments can be considered as unique landmarks. Thus, from a `place descriptor' and localization perspective, these segments do not necessarily need to be interpretable as ``objects''.
However, a standalone image segment descriptor $\textbf{h}_i$ might suffer from \emph{perceptual aliasing} during the localizaton phase. To alleviate this, we add more \emph{place} context to a node from its neighborhood by aggregating descriptors through multi-layered graph convolutions.
This is achieved by simplifying the standard graph convolution network~\cite{kipf2016semi} to compute average node descriptors as below:
\begin{equation}
\mathbf{H}^{(l+1)} = \tilde{\mathbf{D}}^{-1} \tilde{\mathbf{A}} \mathbf{H}^{(l)} \mathbf{I}
    \label{eq:graphConv}
\end{equation}
where $\mathbf{H}$ is the node descriptor matrix (composed of $\textbf{h}$), $A$ is the adjacency matrix for $\mathcal{G}$, $\tilde{\mathbf{A}}=\mathbf{A}+\mathbf{I}$ is the adjacency matrix with self-loops, $\mathbf{I}$ is the identity matrix and $\tilde{\mathbf{D}}$ is the degree matrix for $\tilde{\mathbf{A}}$.
Here, aggregation over successive layers influences a node descriptor through the neighbors of its neighbors, thus gradually expanding the `place' context of any given node. We perform this aggregation on both the map and the query image using $l_{max}=2$.

\subsection{Localization}
In our proposed map with segments-as-nodes, we define localization at the node level through node retrieval. For each of the segment descriptors in the query image, we match it with all the segment nodes in the map and consider it localized if its similarity is greater than a threshold. Although more sophisticated retrieval methods are available, we found that the richness of the descriptor, together with a simple threshold, provided high-quality retrieval. These segment descriptors are informed by their neighbours (see Eq.~\ref{eq:graphConv}), which improves their localization ability due to the added `place' context.

\subsection{Global Planning}
Through the interconnectivity of segments, we aim to obtain navigation plans from our map in the form of \textit{segment tracks} with continuous \textit{hopping} from one track to another, as these segments exit and enter the field of view.
\subsubsection{Edge Weighting}
Given the source and destination segment nodes in our proposed map, we generate a plan using Dijkstra algorithm, where the edge weights 
are set to $0$ and $1$ respectively for inter- and intra-image edges.
This specific design choice is what encourages the \textit{shortest path} search to always prefer edge connections \emph{across} images. It leads to the emergence of \emph{segment tracks} of \textit{persistent entities} that a robot can use as navigation sub-goals, where continuous hopping across the sub-goals of the navigation plan leads to the final destination. We use these edge weights only for generating navigation plans, not for node descriptor aggregation.

\begin{figure*}
    \centering
    \includegraphics[width=\textwidth]{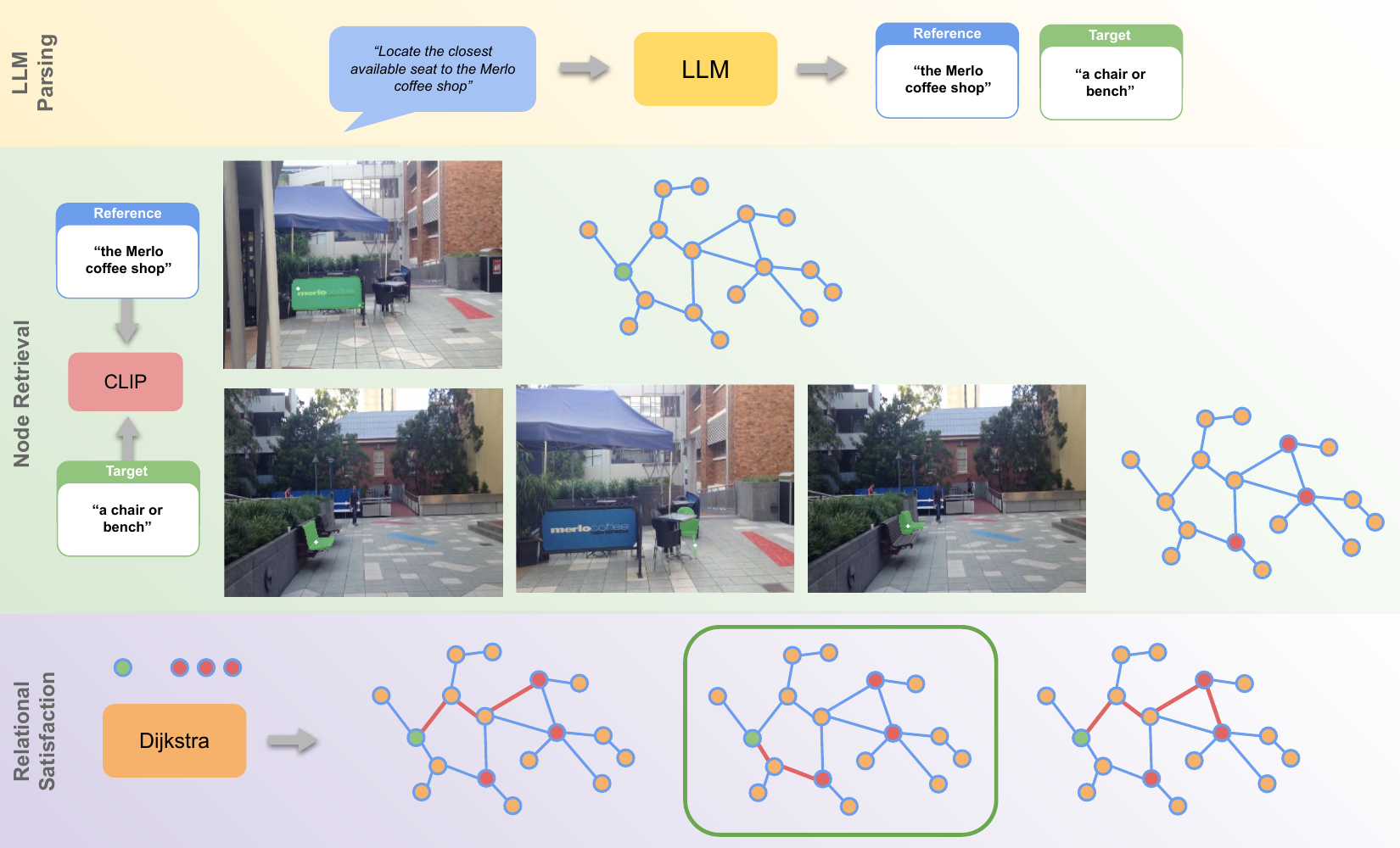}    
    \caption{\textit{Target Object Search Based on Relational Natural Language Queries}: The LLM parses a relational query into a \textit{reference} and \textit{target} node textual description suitable for CLIP to process into language feature vectors. We then retrieve top-3 candidate \textit{target} and \textit{reference} nodes from the map by respectively matching the CLIP language feature vector with the CLIP vision feature vector of each node. Within the topological graph of our map, Dijkstra's algorithm finally selects the object goal for navigation based on the shortest path between the candidate \textit{target} and \textit{reference} nodes.}
    \label{fig:relational}
\end{figure*}

\subsubsection{Planning Strategy}
There exist many different methods~\cite{SPTM2018semi,shah2022lmnav,huang23vlmaps,li2020learning,meng2020scaling} for local motion control that operate on the pair of current observation and sub-goal to generate actions. Since the exact form of input to such controllers, as well as the exact end-task specifications can potentially vary~\cite{li2020learning,wasserman2023last,anderson2018evaluation,wang2022towards}, we define two variants of segment-level plan generation depending on how the intra-image edges are connected. The default mode is to use Delaunay Triangulation (as described in Section~\ref{subsec:SegmentMap}), which we refer to as \textit{Intra-DT} for planning purposes. With intra-image edge weights as $1$, this mode will only ever traverse multiple intra-image neighboring segments when it is able to reach a node that has long inter-image tracks, thus saving the overall path cost. This type of planning can be directly useful for `smooth' robot control as there are no intra-image `long hops'. We also consider an alternative mode of planning, dubbed \textit{Intra-All}, where we create a complete subgraph using all the segments \textit{within} a single image, thus allowing long intra-image hops. This mode of planning can be useful when there is a large number of objects in a single image (e.g., a shelf full of items) which will otherwise incur a high cost for moving from one corner of the image to another. In Section~\ref{sec:results_planning}, we show how these different planning strategies lead to variations in the choice of persistent segment tracks.    

\subsection{Navigation}
We propose two object-level control methods: discrete and continuous, as detailed below. 

\subsubsection{Discrete Control Mode}
For each node in the plan, we match its segment descriptor with all the segment descriptors in the current robot observation (query). The similarity value of the best match determines whether the robot is in the `lost state' (i.e., unable to localize with respect to the reference node, thus explore randomly) or `track state'. For the latter case, we use the horizontal pixel offset of the best matching query segment from the image center to drive the robot towards that object. We use the segment size ratio between the tracked object and its reference to determine a `hop state'. This state implies that the robot has successfully tracked and reached to the reference sub-goal, and can hop on to the next node in the plan and repeat the process until it reaches the last node in the plan. 

\subsubsection{Continuous Control Mode}
In this mode, we use all the segments of the current observation to obtain a control signal. We match all the query segments against all the segments in the local submap (obtained as a set of images within a temporal window of the localized map image). The best matched submap segment corresponding to each query segment is used as a source node to compute path length. These path lengths are used to compute a weighted average of the horizontal pixel offset, thus guiding the robot towards the objects which are closer to the goal. This process is repeated until the minimum path length across matched submap segments reduces to 0. An example of this mode of navigation is shown in Figure~\ref{fig:contControl}.

\subsection{Querying the Map with Open Vocabulary}
We demonstrate one potential use case of our map representation for object-goal navigation based on object-level \textit{relational} queries. We associate each node in our map with a CLIP descriptor of the corresponding image segment, thereby offering an interface for open-vocabulary, natural language queries entailing vague and complex task instructions. More importantly, we introduce an algorithm 
(see Figure~\ref{fig:relational})
that enables generating path plans from complex \textit{relational} queries, e.g., \textit{``locate the closest available seat to the Merlo's coffee shop"}, which exploits the map's ability to capture both intra- and inter-image spatial relationships not present in existing methods. The key here is to identify the \textit{target} (``chairs or benches") and the \textit{reference} (to that target, i.e., ``the Merlo coffee shop") nodes in the scene based on the relational query. We do this by utilising an LLM appropriately prompted to parse the query and identify textual descriptions of these nodes-of-interest. This does not require the LLM to be aware of the map. Across all experiments in the work, we leverage \texttt{GPT-4} as the underlying LLM. The parsed text descriptions of \textit{reference} and \textit{target} are processed into language feature vectors by CLIP's text encoder. We then retrieve top-3 candidate \textit{target} and \textit{reference} nodes from the map by respectively matching the CLIP language feature vector with the CLIP vision feature vector of each node. Within our topological graph, Dijkstra's algorithm finally selects the object goal for navigation based on the shortest path between the candidate \textit{target} and \textit{reference} nodes.

\section{Experiments and Results}
This section details our experimental design and results, aimed at validating the proposed topological map representation for segment-level topological localization, planning for `hopping' based navigation, and object-level control\footnote{Additional implementation details for image preprocessing and models (i.e., SAM~\cite{kirillov2023segment}, DINO~\cite{caron2021emerging}), and CLIP~\cite{radford2021learning}) are in the supplementary.}.

\subsection{Segment-Level Data Association}\label{subsec:data-association}
As the quality of segment-level data association lies at the heart of the robustness and integrity of our mapping, as well as for the plans made within these maps, we conduct experiments to evaluate the efficacy of the data association component of our pipeline. Our method is simple but backed by rich descriptors based on local and broader contextual information.
We consider two kinds of experiments on real-world data, which are outlined in more detail below.
In the first set of experiments, the ground truth segments and instances are available indoors, such as GibsonEnv~\cite{xiazamirhe2018gibsonenv}, This availability allows us to perform quantitative evaluation of segment-level association. However, in the second set of experiments, the lack of similar ground truth data outdoors means that we must resort to evaluating a downstream task -- localisation -- to assess its performance based on our segment correspondences.

\begin{table}
    \caption{Accuracy of segment-level object recognition.
    }
    \centering
    \resizebox{0.4\textwidth}{!}{%
    \begin{tabular}{ccc}
        \toprule
        \textbf{} & \textbf{CLIP~\cite{radford2021learning}} & \textbf{DINO~\cite{caron2021emerging}} \\
        \midrule
        \textbf{Object Instance Recognition} & 35.11\% & 56.43\% \\
        \textbf{Object Category Recognition} & 62.87\% & 79.04\% \\
        \bottomrule
    \end{tabular}
    }
    \label{tab:objInstaCatRec}
\end{table}

\def\scaleMatchingImg{0.13}
\begin{figure}
\setlength{\tabcolsep}{1pt}

    \centering
    \begin{tabular}{ccccc}

    \rotatebox{90}{\hspace{0.2cm}\scriptsize Query}&
        \includegraphics[scale=\scaleMatchingImg]{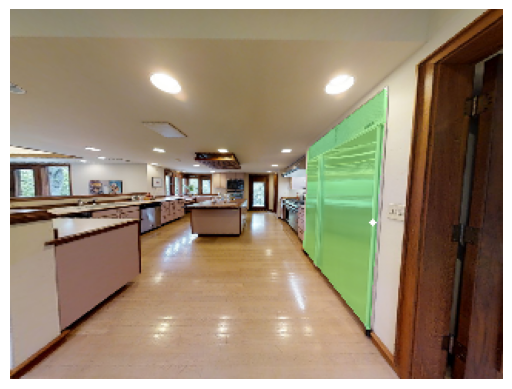}\hphantom{\checkmark} &
\includegraphics[scale=\scaleMatchingImg]{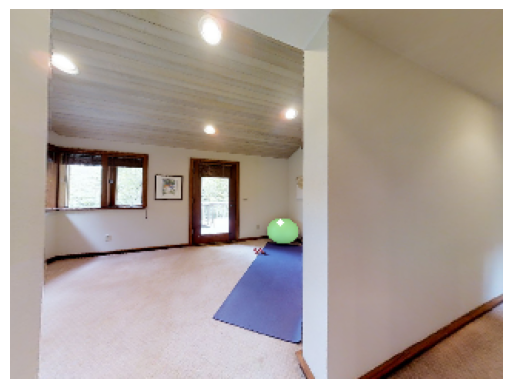}\hphantom{\checkmark} &
        \includegraphics[scale=\scaleMatchingImg]{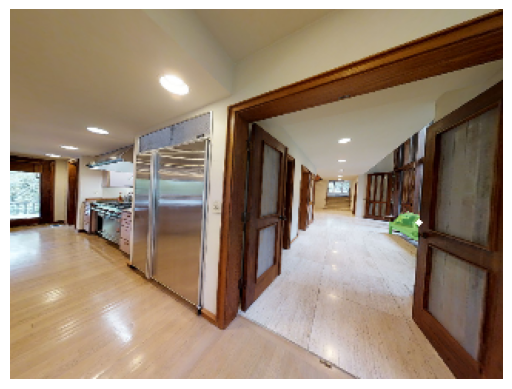}\hphantom{\checkmark} &
        \includegraphics[scale=\scaleMatchingImg]{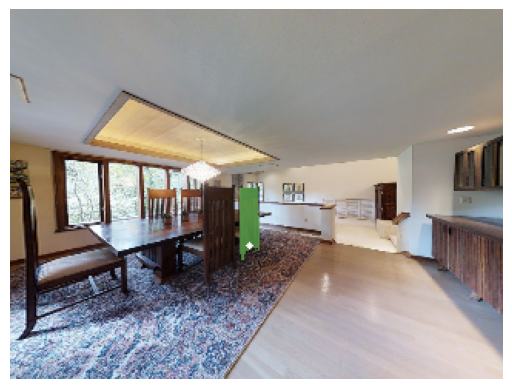}\hphantom{\checkmark} \\

            \rotatebox{90}{\hspace{0.2cm}\scriptsize DINO}&
        \includegraphics[scale=\scaleMatchingImg]{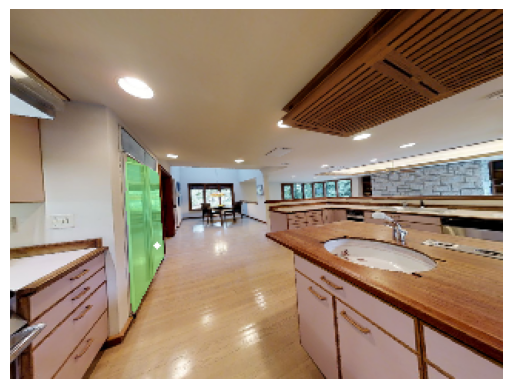}\checkmark &
        \includegraphics[scale=\scaleMatchingImg]{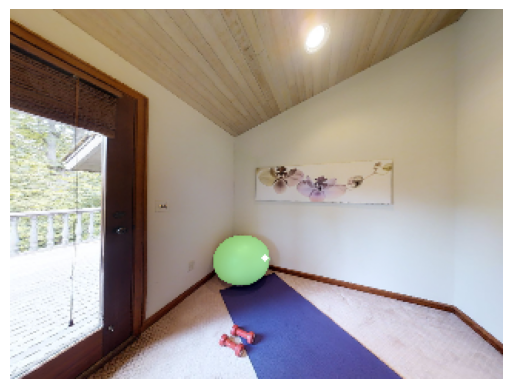}\checkmark &
        \includegraphics[scale=\scaleMatchingImg]{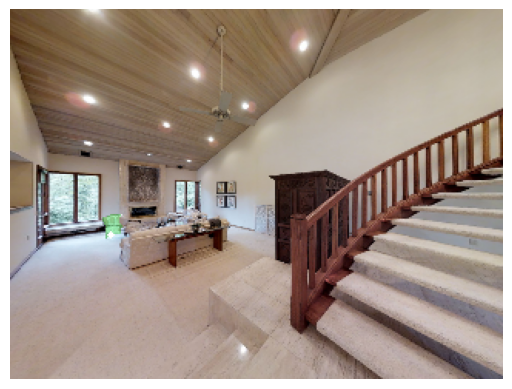}\texttimes &
        \includegraphics[scale=\scaleMatchingImg]{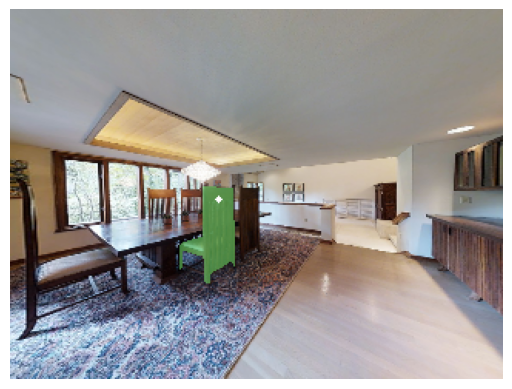}\hphantom{\checkmark}\\%\texttimes\checkmark &

    \rotatebox{90}{\hspace{0.2cm}\scriptsize CLIP}&
        \includegraphics[scale=\scaleMatchingImg]{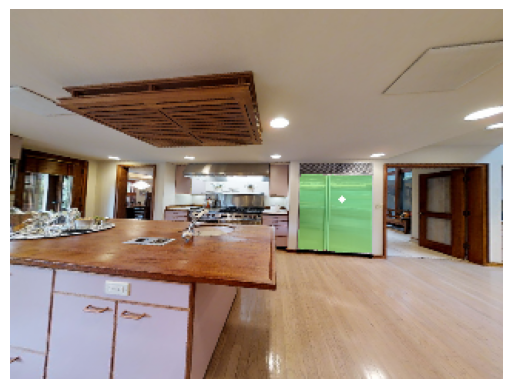}\checkmark &
        \includegraphics[scale=\scaleMatchingImg]{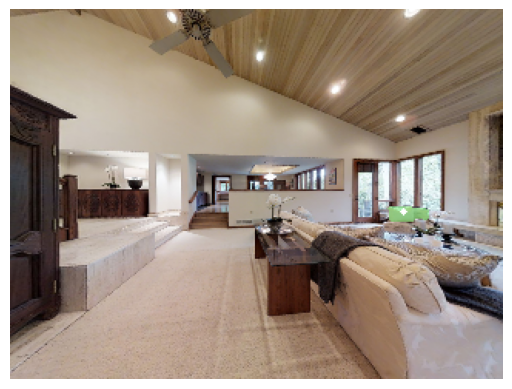}\texttimes &
        \includegraphics[scale=\scaleMatchingImg]{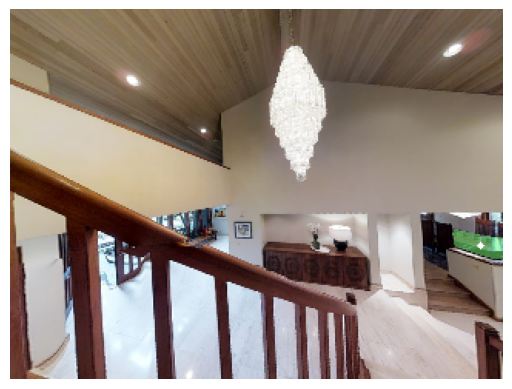}\texttimes &
        \includegraphics[scale=\scaleMatchingImg]{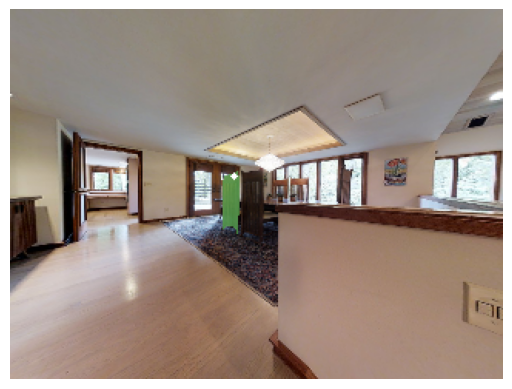}\hphantom{\checkmark}%

    \end{tabular}
    \caption{Object Instance Recognition in GibsonEnv~\cite{xiazamirhe2018gibsonenv}: The rows show segment masks (in green) for the query, DINO match, and CLIP match respectively. Symbols (\checkmark/\texttimes) adjacent to images indicate success or failure in association. The final column illustrates category-level recognition success despite both methods failing at the instance level (multiple chairs in close proximity).}
    \label{fig:recogn_examples}
\end{figure}

\subsubsection{Object Instance and Category Recognition}
In this experiment, to demonstrate the efficacy of our segment-level association, we make use of ground truth detections and segmentation of instances in an indoor environment: GibsonEnv~\cite{xiazamirhe2018gibsonenv}. In particular, we show here examples from the house \texttt{Klickitat} as it is representative of the diverse range of environments in the dataset. 
To align with the standard input requirements of SAM, and to ``simulate" a forward-facing camera, we extract perspective images with a field-of-view of 120 degrees from the real-world GibsonEnv panoramas and treat these as the raw images.
Next, we obtain class-agnostic SAM segments from each image and assign these segments to their corresponding ground truth object instances in each image using Intersection over Union (IoU), with a minimum threshold of $0.2$. 
To ensure data quality, we consistently exclude segments with sizes comprising less than $0.2\%$ of the overall image.
Finally, for this experiment, we have a total of $544$ distinct views (SAM segments) of $68$ unique objects from $18$ diverse categories.
We assess the quality of descriptors (such as DINO~\cite{caron2021emerging} and CLIP~\cite{radford2021learning}) for segment-level association by evaluating the (top-1) accuracy of our descriptor matching with the correct object. As explained in Section~\ref{subsec:SegmentMap}, the matches are selected based on the nearest neighbour criterion over descriptors.

Table~\ref{tab:objInstaCatRec} shows a comparative analysis of different descriptors for object instance and category recognition from diverse viewpoints.
It is apparent that DINO achieves better results than CLIP in this context, which can be attributed to differences in how they are supervised and their training objectives.  
While CLIP performs reasonably well in predicting categories, DINO features exhibit greater distinctiveness in both instance-level and category-level recognition.
In Figure~\ref{fig:recogn_examples}, we show some of the object instance and category recognition outcomes, featuring both successful and unsuccessful cases.

\subsubsection{Segment-level Topological Localization}
\begin{figure}
    \centering
    \begin{tabular}{cc}
    \includegraphics[width=0.2\textwidth]{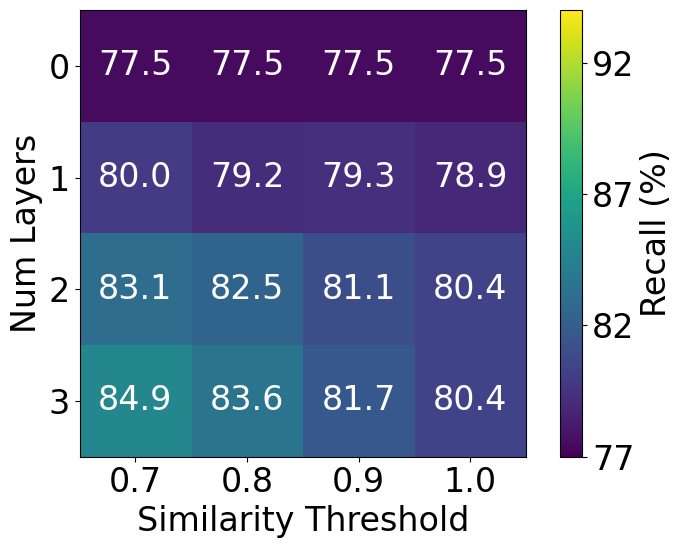} &
    \includegraphics[width=0.2\textwidth]{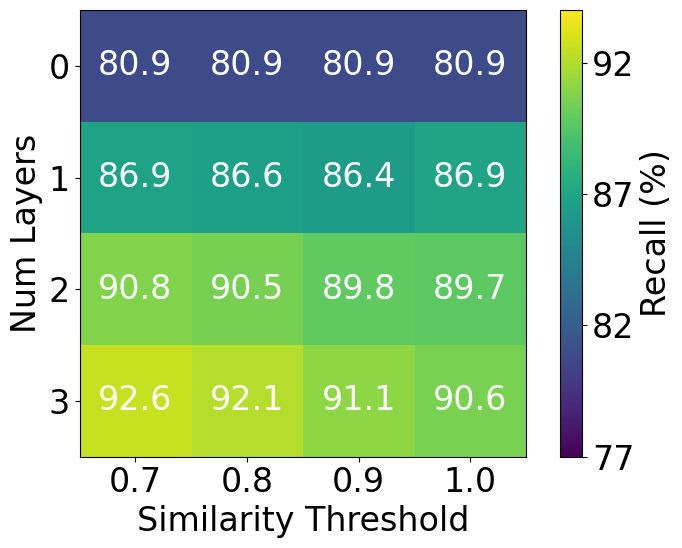}
    \end{tabular}
    \caption{Node-level localization across varying number of graph convolutional layers (y-axis) and incremental inclusion of inter-image edges based on a similarity threshold (x-axis) for DINO (left) and DINOv2 (right).}
    \label{fig:nodeRecGraph}
\end{figure}

Since segment- or object instance-level ground truth associations are not always available, we also conduct experiments to measure the quality of both our map and the localization ability through a segment-level topological localization task. For this purpose, we use a popular visual place recognition dataset, GPCampus~\cite{arren_glover_2014_4590133}, which comprises three traverses of a University Campus: two day and one night time. We only use its Day Left and Day Right traverse as the reference map and query set respectively. We coarsely evaluate segment-level association by first tagging both the query segment and its matched segment to their respective image indices, and then using these associated images to compute Recall@1 based on a localization radius of $5$ frames. Figure~\ref{fig:nodeRecGraph} shows that segment-level recognition for both DINO (left) and DINOv2 (right) improves with an increasing number of graph convolution layers as well as incremental inclusion of inter-image edges. The former only considers segments from within an image while the latter resembles sequential descriptor-type place recognition~\cite{garg2021seqnet}. 

\subsection{Planning}\label{sec:results_planning}
We show qualitative results of our full pipeline using two complementary datasets. a) \textit{PanoContext-Living}, which refers to one of the living room panoramic images (\texttt{2cfc836333}) from the original PanoContext dataset~\cite{zhang2014panocontext,zou2018layoutnet}. We split this pano image uniformly along the horizontal axis to create multiple frames, with a horizontal wraparound. Thus, this dataset represents a pure rotation-based robot traversal. We explicitly compute data association between the last and the first frame to close the loop. b) \textit{GPCampus-DayLeft}~\cite{arren_glover_2014_4590133}, which is a forward-moving robot traverse. 
For both these datasets, we first construct the segment-level map, then query the resultant graph with text to identify source and target node based on CLIP similarity, and then finally generate a plan between these pairs of nodes. 

\def\scalePanoContext{0.2\textwidth}

\begin{figure}
    \centering
\begin{tabular}{cc}
    Intra-All  & Intra-DT \\

    \includegraphics[width=\scalePanoContext]{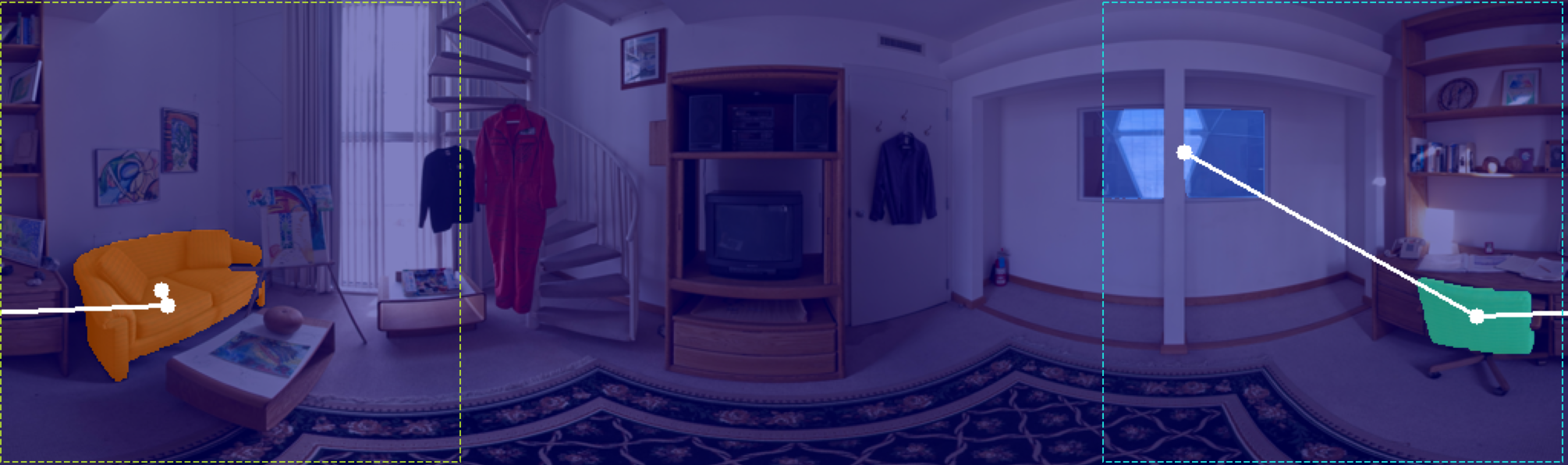} 
    &
    \includegraphics[width=\scalePanoContext]{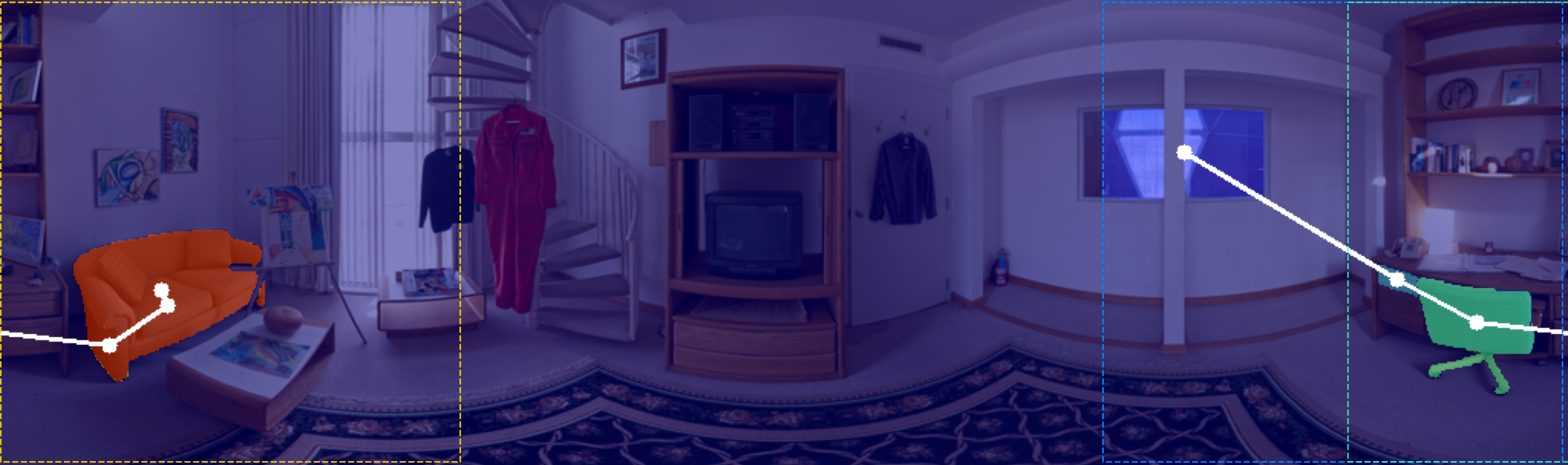} 
    \\
    \multicolumn{2}{c}{(a) \texttt{Window} to \texttt{Sofa}}\\
    
    \includegraphics[width=\scalePanoContext]{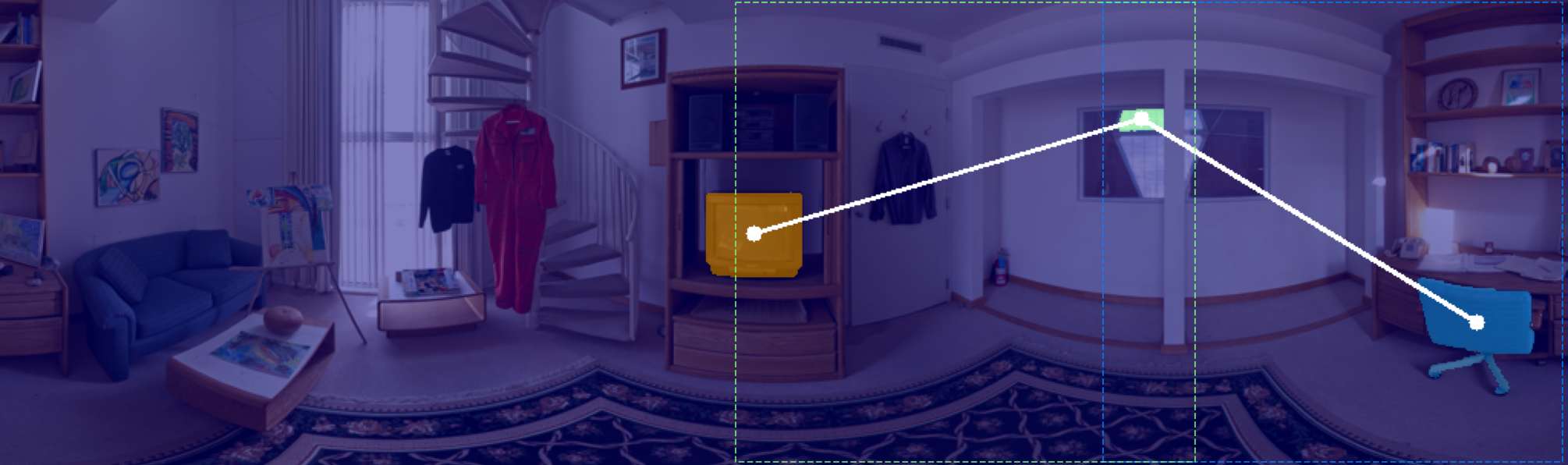} 
    &
    \includegraphics[width=\scalePanoContext]{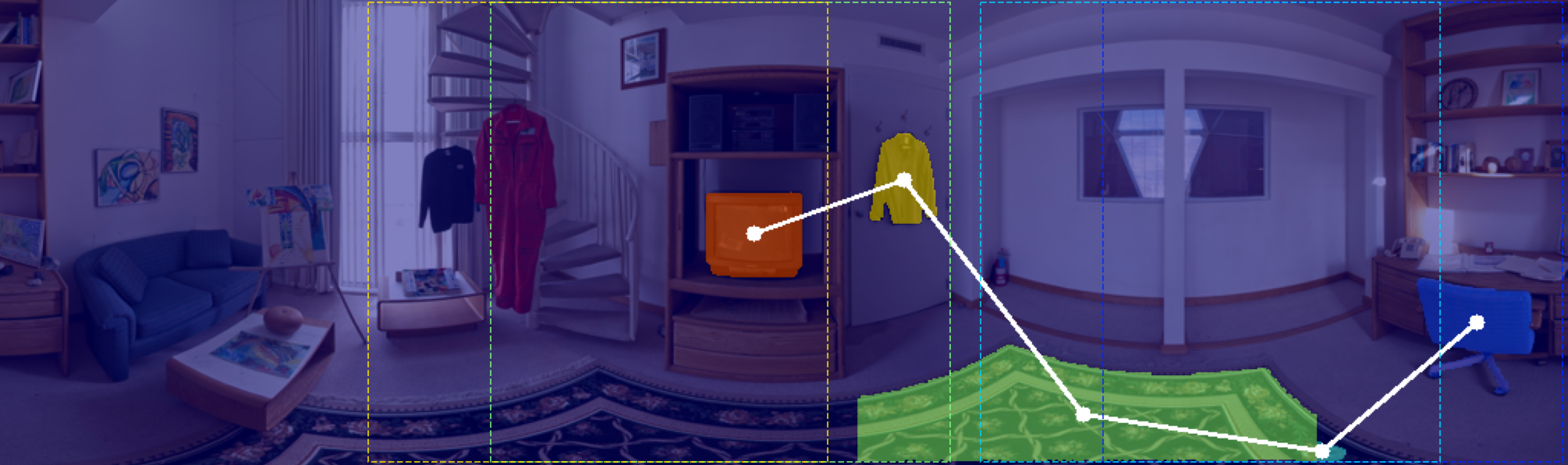}  
    \\
    \multicolumn{2}{c}{(b) \texttt{Chair with wheels} to \texttt{Television}}\\

    \includegraphics[width=\scalePanoContext]{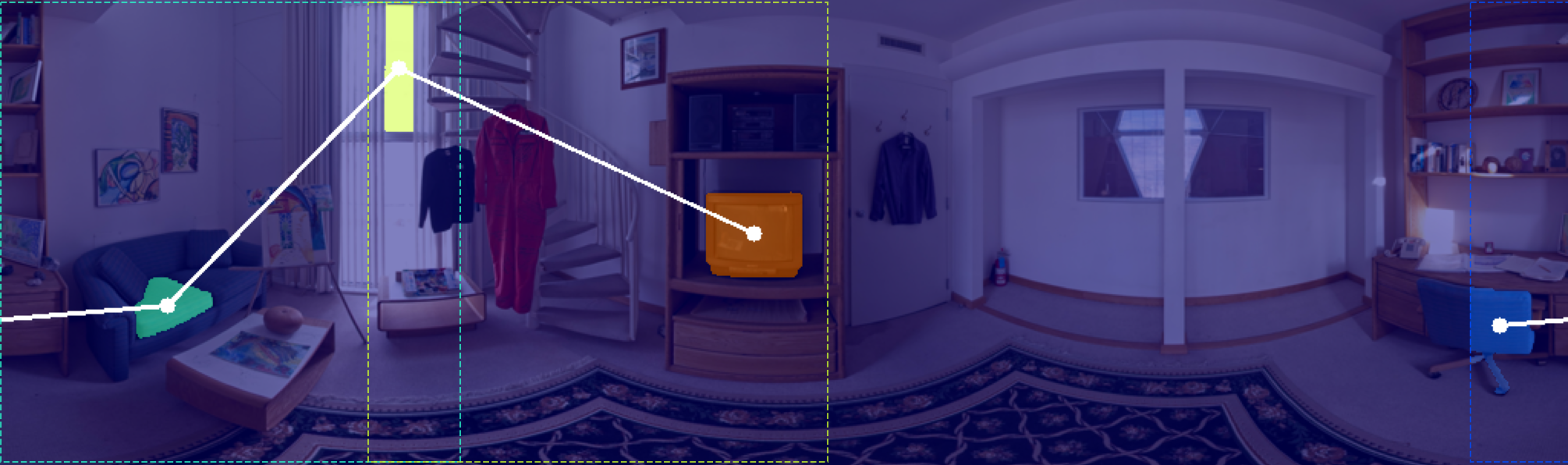} 
    &
    \includegraphics[width=\scalePanoContext]{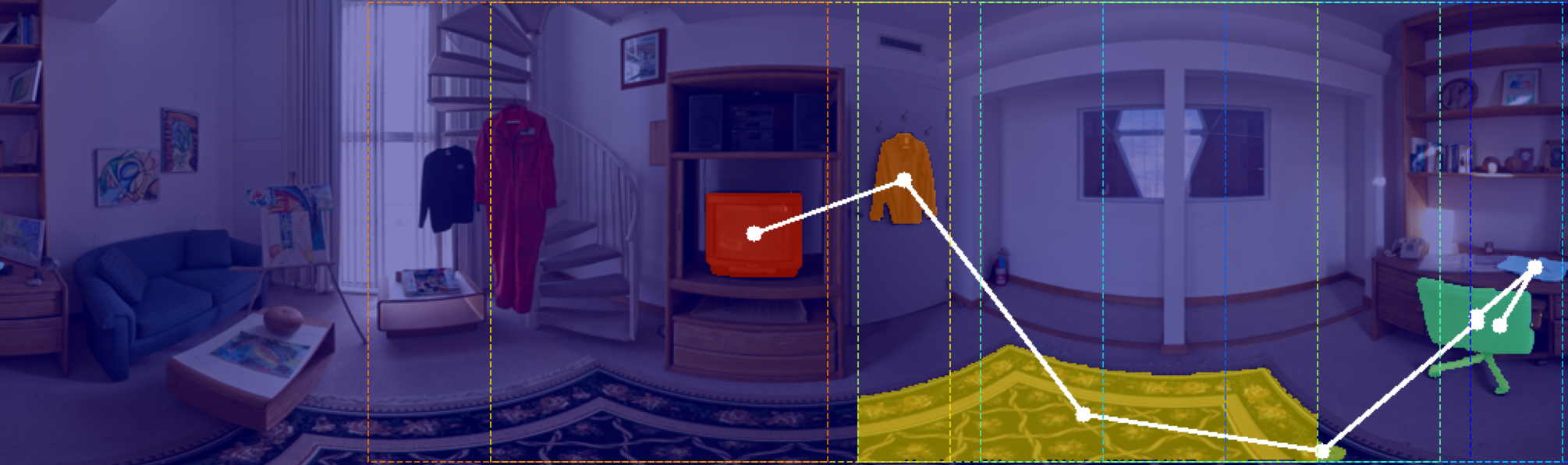} 
    \\
    \multicolumn{2}{c}{(c) \texttt{Chair} to \texttt{Television}}\\
    
    \end{tabular}
    \caption{Segment-level plans using text queries for source and target, showing shortest paths for panoramic `pure rotation'.}
    \label{fig:layout_panoSplit_plans}
\end{figure}

\subsubsection{PanoContext-Living}
Figure~\ref{fig:layout_panoSplit_plans} shows multiple plans using a variety of text queries for both types of planning strategies: Intra-All and Intra-DT. Each of the selected segments and their connectivity based on the shortest path is shown, with path edges wrapped around the pano image. The subsampled frames from the pano are shown as dashed boxes in color corresponding to the segment belonging to that frame. 

\paragraph*{Intra-All} For Intra-All planning on this pure rotation setting, the inferred shortest path can be coarsely related to the horizontal offset (allowing wraparound) between the pixel centroids of the source and the target segment. In Figure~\ref{fig:layout_panoSplit_plans}(a) (Intra-All), for text queries \texttt{Window} (source) and \texttt{Sofa} (target), the shortest path is correctly found from the wraparound frames via \texttt{Chair}. In examples (b) and (c), we extract paths to \texttt{Television} from \texttt{Chair with wheels} and \texttt{Chair}. Indicating imperfections of the SAM+CLIP combination, \texttt{Chair} finds the best match with one of its partial visual observation, in contrast to \texttt{Chair with wheels} which matches correctly with the full chair. Nevertheless, both the paths in (b) and (c) are practically similar in terms of the number of yaw steps needed to reach the target. 

\paragraph*{Intra-DT} For the Intra-DT plans, in all the cases, paths span multiple objects (more than the Intra-All), inducing a smoother transition from source to target. In examples (b) and (c), the paths are composed of the carpet nodes -- this consistent choice is justified from an almost `omnipresence' of carpet throughout the scene, as it had not been filtered out in our preprocessing of common segments. Thus, in both the cases, intra-image hops try to land on to the carpet node to reach the target with the least inferred cost.

\def\scaleImgPlan{0.13}
\begin{figure}
\setlength{\tabcolsep}{1pt}
    \centering
    \begin{tabular}{cccccc}

    \rotatebox{90}{\hspace{0.05cm}\scriptsize Intra-DT}&
    \includegraphics[scale=\scaleImgPlan,trim={0 2.5cm 1cm 0},clip]{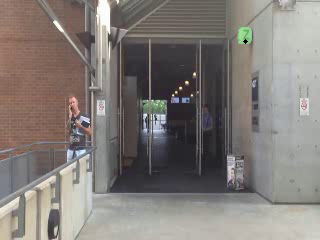} &
    \includegraphics[scale=\scaleImgPlan,trim={0 2.5cm 1cm 0},clip]{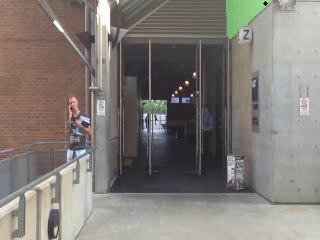} &
    \includegraphics[scale=\scaleImgPlan,trim={0 2.5cm 1cm 0},clip]{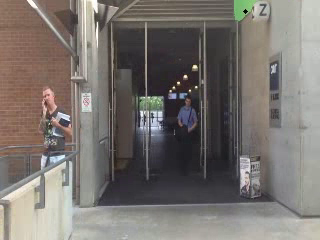} &
    \includegraphics[scale=\scaleImgPlan,trim={0 2.5cm 1cm 0},clip]{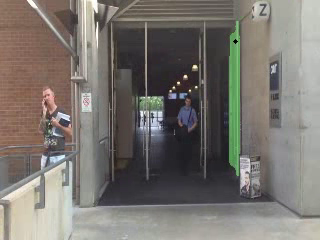} &
    \includegraphics[scale=\scaleImgPlan,trim={0 2.5cm 1cm 0},clip]{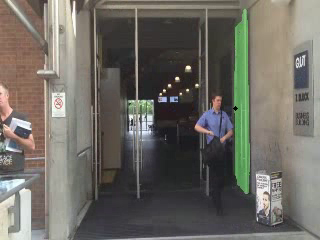} \\

    \rotatebox{90}{\hspace{0.05cm}\scriptsize Intra-All}&
    \includegraphics[scale=\scaleImgPlan,trim={0 2.5cm 1cm 0},clip]{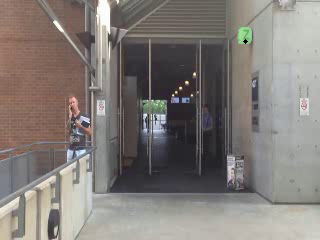} &
    \includegraphics[scale=\scaleImgPlan,trim={0 2.5cm 1cm 0},clip]{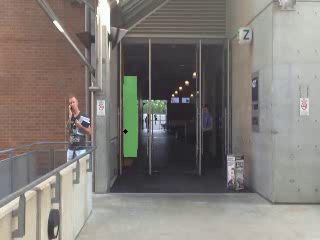} &
    \includegraphics[scale=\scaleImgPlan,trim={0 2.5cm 1cm 0},clip]{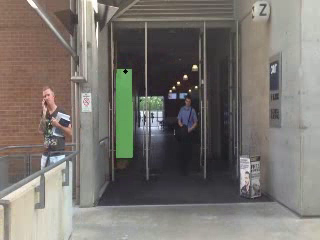} &
    \includegraphics[scale=\scaleImgPlan,trim={0 2.5cm 1cm 0},clip]{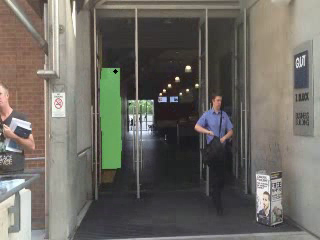} &
    \includegraphics[scale=\scaleImgPlan,trim={0 2.5cm 1cm 0},clip]{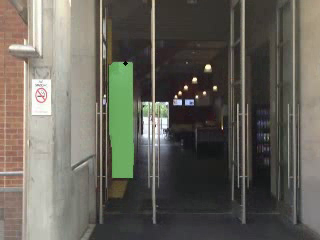} \\

    \rotatebox{90}{\hspace{0.05cm}\scriptsize DA-All}&
    \includegraphics[scale=\scaleImgPlan,trim={0 2.5cm 1cm 0},clip]{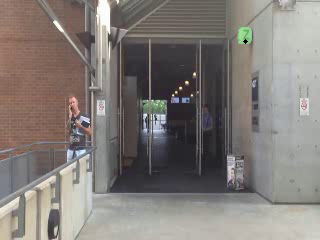} &
    \includegraphics[scale=\scaleImgPlan,trim={0 2.5cm 1cm 0},clip]{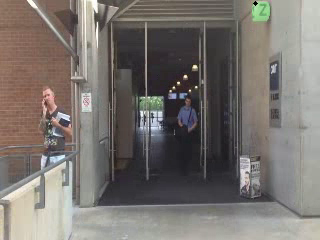} &
    \includegraphics[scale=\scaleImgPlan,trim={0 2.5cm 1cm 0},clip]{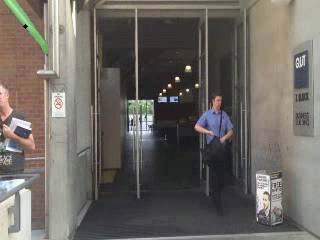} &
    \includegraphics[scale=\scaleImgPlan,trim={0 2.5cm 1cm 0},clip]{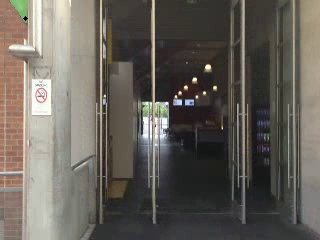} &
    \includegraphics[scale=\scaleImgPlan,trim={0 2.5cm 1cm 0},clip]{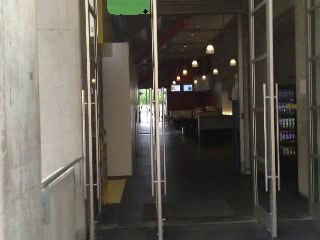}

    \end{tabular}
    \caption{Variations in segment-level navigation plans (one per column) depending on how the edges are defined and weighted for path search.}
    \label{fig:plans}
\end{figure}

\subsubsection{GPCampus-DayLeft}
In Figure~\ref{fig:plans}, we show the segment-level plan for the forward-moving robot traverse, with \texttt{Z block} and \texttt{Dustbin} as  the source and target text queries. Here, we only show the planned segments close to the source node, please refer to the supplementary video for the full plan visualization.
The first two rows correspond to the Intra-DT and Intra-All planning, and the last row corresponds to a naive baseline where an inter-image edge for each of the segments is included without any similarity thresholding (see Eq.~\ref{eq:edge}). This implies that during planning there always exists a $0$ cost inter-image edge for all the segments, thus never needing to traverse an intra-image edge. 
In the Intra-DT row, the first $4$ frames (columns) show an intra-image traversal to reach the \texttt{door} which has a persistent track over multiple frames. In the Intra-All row, it can be observed that a single intra-image hop directly leads to a persistent track of a \texttt{closet}. In the DA-All row, the paths are formed based on rapid hopping, as soon as the current tracked object goes out of the field-of-view, regardless of any persistent segment tracks.

\def\scaleCC{0.175}
\begin{figure}
\setlength{\tabcolsep}{1pt}
    \centering
    \begin{tabular}{cccc}
    \includegraphics[scale=\scaleCC]{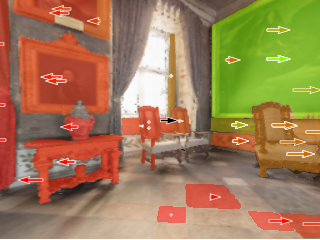} &
\includegraphics[scale=\scaleCC]{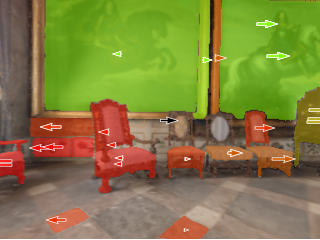} &
\includegraphics[scale=\scaleCC]{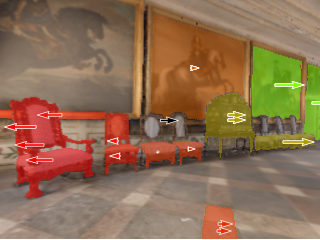} &
\includegraphics[scale=\scaleCC]{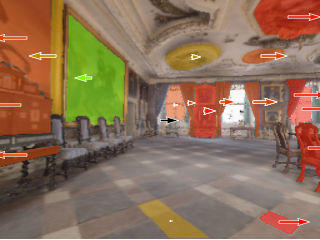}
    \end{tabular}
    \caption{Successful navigation example in Habitat using continuous control mode to reach the green painting goal in the rightmost image. The horizontal pixel offset (depicted through the length and direction of arrow) for each of the matched query segments is weighted by the path length to the goal (depicted through color with length decreasing from red to green), to generate an aggregated angular velocity.}
    \label{fig:contControl}
\end{figure}

\subsection{Navigation}
We conducted preliminary trials of zero-shot robot navigation using segment-level mapping and planning, both in real world and simulation. We initialize the robot pose such that the first reference map node (sub-goal) of the plan is in its field of view. We use PID controller to convert the horizontal pixel offset into yaw velocity, while the forward translation is always fixed to a small velocity. Figure~\ref{fig:contControl} shows an example of continuous control mode in Habitat simulator~\cite{savva2019habitat}. We defined an initial trajectory in its \texttt{skokloster} environment by sampling multiple farthest navigable points. At inference, the robot was then tasked to go from one of the random points along the trajectory to another. Our trials (in supplementary video) show that our proposed representation, powered by the foundation models SAM and DINO, enables embodiment-agnostic control strategies for zero-shot goal-directed navigation without needing to train data-hungry task-specific policies.

\section{Limitations}
While our approach exhibits notable strengths in segment-level topological mapping and planning for spatial reasoning and navigation, it also has multiple limitations worth discussing. 
\textit{a)} The efficacy of our approach is strongly tied to the quality of segment-level data association. We observed failures in navigation trials due to mismatches caused by repetitive structures. We found LightGlue~\cite{lindenberger2023lightglue} to perform better than DINOv2 for segment association in highly aliased environments (e.g., paintings and chairs in Figure~\ref{fig:contControl}). \textit{b)} Our method in its current form cannot deal with dynamic changes in the environment. \textit{c)} Considering `things' vs `stuff', despite the convenience of semantic preprocessing enabled by the combination of SAM and CLIP to remove `stuff', some segments from ground or walls can still persist. \textit{d)} In our navigation experiments, we found that the lack of repeatable segmentation during the revisits led to incorrect area ratio, thus affecting the forward/backward motion and `hop state' decision -- this could though be addressed through depth information (used solely for this purpose, while still using the topological map). \textit{e)} Finally, we note that handling relational queries through LLMs is prone to failures in cases where metric information is necessary to deem two objects being next to each other.

\section{Conclusion and Future Work}
This paper presented a novel topological map representation centred on \textit{image segments}, which serve as semantically-rich, open-vocabulary queryable nodes within a topological graph. The method uses an integrated strategy involving segment-level data association and segment-level planning for object-goal navigation. Our preliminary trials on segment-level \textit{hopping} based navigation indicate that powerful foundation models like SAM (for segmentation) and DINOv2 (for data association) can enable zero-shot navigation without requiring 3D maps, image poses or a learnt policy.

There are several promising directions for future work. One avenue involves incorporating visual servoing-based navigation to provide real-time visual feedback, which could improve the system's navigation capabilities and robustness. Furthermore, while our current approach predominantly relies on topological mapping, integrating \textit{local} node- and edge-level metric information can introduce a higher degree of granularity and precision, thereby enhancing the system's navigation capabilities. Finally, semantically labelling each node could facilitate the construction of 3D scene graph representations suitable for higher-level task planning~\cite{rana2023sayplan}.

\bibliographystyle{IEEEtran}
\bibliography{references}

\end{document}